\documentclass[runningheads]{llncs}

\usepackage[T1]{fontenc}
\usepackage{graphicx,verbatim}
\usepackage{amsmath,amssymb,amsfonts}
\usepackage[ruled,vlined,linesnumbered]{algorithm2e}
\usepackage{booktabs}
\usepackage{xcolor}
\usepackage{hyperref}
\usepackage{enumitem}
\graphicspath{{figs/}{figures/}}

\hypersetup{colorlinks=true,linkcolor=blue!60!black,citecolor=green!50!black,urlcolor=blue!70!black}

\newcommand{\R}{\mathbb{R}}

\newcommand{\N}{\mathcal{N}}
\newcommand{\bx}{\mathbf{x}}

\newcommand{\bmu}{\boldsymbol{\mu}}
\newcommand{\bSigma}{\boldsymbol{\Sigma}}

\begin{document}

\title{Gaussian Meta-Space Augmentation for Stacking Ensembles in Multimodal IPMN Risk Stratification}
\titlerunning{Augmenting Stacking Ensembles for IPMN Risk}
\author{%
Max A. Nelson\inst{1} \and
Eminenur Sen Tasci\inst{1} \and
Zhixiang Wang\inst{1} \and
Zongwei Zhou\inst{2} \and
Halil Ertugrul Aktas\inst{1} \and
Andrea M. Bejar\inst{1} \and
Elif Keles\inst{1} \and
Ziliang Hong\inst{1} \and
S{\i}tk{\i} Safa Taflan\inst{3} \and
Muhammed Enes Tasci\inst{1} \and
Frank H. Miller\inst{1} \and
Michael B. Wallace\inst{4} \and
Rajesh N. Keswani\inst{5} \and
Gorkem Durak\inst{1} \and
Ulas Bagci\inst{1}}
\authorrunning{M. A. Nelson et al.}
\institute{Machine \& Hybrid Intelligence Lab, Department of Radiology, Northwestern University, Chicago, IL, USA \and
Department of Computer Science, Johns Hopkins University, Baltimore, MD, USA \and
Istanbul Faculty of Medicine, Istanbul University, Istanbul, Turkey \and
Division of Gastroenterology and Hepatology, Mayo Clinic Florida, Jacksonville, FL, USA \and
Department of Gastroenterology and Hepatology, Northwestern University, Chicago, IL, USA \\
\email{max.nelson1@northwestern.edu}}
\maketitle

\begin{abstract}

Pancreatic cancer is among the most lethal malignancies; risk
stratification of intraductal papillary mucinous neoplasms (IPMNs) offers a
crucial opportunity for early intervention but typically requires invasive tissue biopsy. Dominant vision-based approaches, including radiomics and deep learning, provide promising but initially separate discrimination opportunities. Similarly, multisequence MRI (T1W/T2W) and anatomically decomposed (head, body and tail) analysis of the pancreas provide additional and potentially complementary signals. Effective fusion of this information is crucial in ordinal IPMN dysplasia risk prediction and can be accomplished via a meticulously regularized and calibrated ensemble stacking combiner. We present \textbf{cUPMI}, a class-conditional Gaussian augmentation of a combiner’s log-probability meta-features and test it on various prediction paradigms. In our multi-center analysis, we find cUPMI adds limited value to properly
regularized $L_2$-logistic binary classification stacks, but consistently regularizes higher-capacity tree combiners in the binary and radiomics-only setting
(RF $\Delta{+}0.015$ and XGBoost $\Delta{+}0.024$ binary AUC, positive in all
seeds). Its cleanest ordinal benefit appears for XGBoost on an 8-stream
Radiomics task (3-class no$<$low$<$high, $+0.022$ QWK in all seeds).
Separately, fold-locked fusion of radiomics and 2.5D CNN streams yields the
strongest overall model, an RF stack reaching QWK $0.595$ ($95\%$ CI
$[0.54,0.64]$) and binary AUC $0.839$, surpassing radiomics, 2.5D
ResNet, and 3D DenseNet-121 baselines.

\end{abstract}

\begin{figure}[t]
\centering
\includegraphics[width=\textwidth]{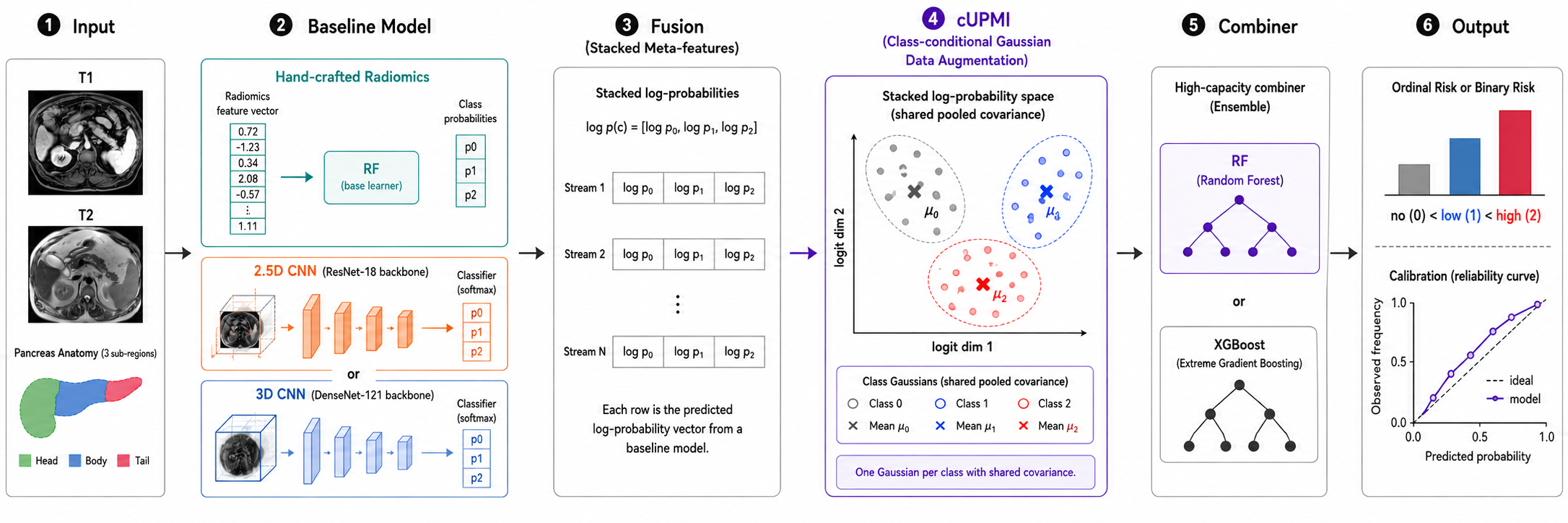}
\caption{Stacked-fusion pipeline. Radiomics and CNN base learners produce per-stream
probabilities, concatenated fold-locked into stacked log-probability meta-features. cUPMI augments
the training fold with class-conditional Gaussian synthetic meta-features upstream of a high-capacity level-1
combiner.}
\label{fig:cupmi_pip}
\end{figure}

\section{Introduction}\label{sec:intro}

Intraductal papillary mucinous neoplasms (IPMNs) are common incidental
findings on imaging \cite{chang2016incidental} and are detected in
$\sim\!10\%$ of individuals aged $\geq 50$ years \cite{delafuente2023ipmn}.
IPMNs arise from the main pancreatic duct (MD-IPMN) and/or its side branches
(BD-IPMN) and carry a considerable risk of malignant transformation
\cite{tanaka2017fukuoka,delchiaro2018european}. Accordingly,
imaging-based assessment is essential to detect features suggestive of
malignancy and to differentiate lesions that warrant surveillance from those
requiring surgical resection. Current management relies on consensus criteria
\cite{ohtsuka2024kyoto} that provide suboptimal discrimination between
higher- and lower-risk IPMN
\cite{qadir2025ai_ipmn_plosdighealth,robles2016icg,vandenbulcke2021guidelines},
motivating more accurate and graded approaches to stratification of
malignancy risk.

Magnetic resonance imaging (MRI) is the preferred modality for characterizing
pancreatic cysts, enabling evaluation of ductal communication and key
morphological findings such as mural nodules and septations
\cite{delchiaro2018european,tanaka2017fukuoka}. T1-weighted (T1W) and
T2-weighted (T2W) sequences are standard for IPMN surveillance
\cite{delchiaro2018european,elta2018acg}. However, visual assessment of these
sequences is variable and may miss subtle imaging patterns associated with
malignant transformation
\cite{miller2022pancreatic_cystic_lesions_field_defect}. More standardized and
quantitative approaches---spanning radiomics and, increasingly, deep
learning---are needed to overcome the limitations of the human eye.

Effectively combining multiple streams of signal remains a central technical challenge, particularly when imaging features are extracted from disjoint anatomical subregions. Patients in our cohort present with IPMN cysts located throughout different and sometimes multiple regions of the pancreas. This motivates production of both confined sub-regional labels (i.e., head/body/tail) and larger whole-organ masks. The stream count increases further when both T1W and T2W sequences are available and when different representations are considered, namely engineered radiomics features and learned convolutional neural network (CNN) descriptors. These sequence, region, and representation choices define distinct streams that must be combined without overfitting the downstream predictor.

Multi-modality fusion techniques attempt to leverage
complementary information from multiple sources and commonly encompass early
fusion (feature concatenation) or decision-level fusion (averaging or
stacking)~\cite{huang2020fusion,stahlschmidt2022multimodal}. Early fusion
poses dimensionality concerns as large
cohorts are often required to mitigate overfitting ~\cite{huang2020fusion} and feature-mix ratios are ostensibly determined arbitrarily. Late
techniques fare better by training stream-specific models and combining their
predictions~\cite{huang2020fusion}, directly circumventing dimensionality expansion. In the context of late fusion, stacked generalization addresses differences in baseline predictive power and noise by training a level-1 combiner over predictions from base models
\cite{wolpert1992stacked,ting1997stacked}. The combiner is learned from a
\emph{small} meta-feature set, and high-capacity combiners (e.g., random
forest, gradient boosting) can readily overfit. Conversely, a well-regularized linear combiner can leave modality interactions unexploited and can be insufficient for multiclass classification tasks \cite{seewald2002stacking}.

We introduce calibrated Upstream Probabilistic Meta-Imputation
(cUPMI), a \emph{regularizer for level-1 combiners}. cUPMI fits one
class-conditional Gaussian per class to the stacking meta-features in
log-probability space, using a shared pooled covariance, and injects synthetic
samples upstream of the combiner to smooth its decision surface. This is a
mixture only across classes, not a BIC-selected within-class mixture, which is a
deliberate choice at this sample size.

Under a leakage-controlled nested protocol, cUPMI gives no advantage over an already-regularized $L_2$-logistic stack for binary discrimination, but consistently regularizes higher-capacity tree combiners on ordinal classification tasks. Separately, a fold-locked cross-track fusion of deep-learning and radiomics streams across whole-organ and sub-regional segmentations provides the strongest overall discrimination, surpassing every single-architecture baseline.

\section{Data and Preprocessing}\label{data}
Our retrospective analysis comprises 678 unique patients (Table~\ref{tab:data-breakdown}) from the multi-center Cyst-X collection \cite{cystx2025dataset}.
MRI sequences were acquired between March 2004 and June 2024 on GE, Siemens, and Philips scanners (1.5T/3T). All patient data were de-identified prior to analysis, and all subjects are adults aged $\geq$18 years. Based on segmentation availability, we construct two paired T1W/T2W analysis cohorts that differ by region of interest. The \emph{whole-organ} paired cohort retains cases with co-registered T1W/T2W sequences and whole-pancreas masks ($n{=}629$ for the 3-class protocol; $n{=}629$ in the binary setting, distributed \{low-risk~487 / high-risk~142\}). The \emph{sub-regional} cohort additionally requires valid head/body/tail part-masks and is analyzed independently per region ($n{=}616$ for 3-class fusion, distributed \{no-risk~154 / low~320 / high~142\}; binary \{474 / 142\}).

Labeling was standardized across centers using a predefined protocol based on histopathology or long-term imaging follow-up. The binary protocol categorized cases as: (i) low-risk, including normal pancreatic imaging without cystic lesions, benign non-IPMN cysts, low- or intermediate-grade dysplasia IPMNs, or lesions stable for at least 3 years without worrisome features or high-risk stigmata; and (ii) high-risk, including IPMNs with high-grade dysplasia, carcinoma in situ, or invasive carcinoma confirmed by EUS-guided biopsy or surgical pathology, with MRI obtained within 6 months of diagnostic confirmation. The 3-class protocol further divided the binary low-risk group into: (i) no-risk, including normal pancreatic imaging or benign non-IPMN cysts; and (ii) low-risk, including low- or intermediate-grade dysplasia IPMNs or lesions stable for at least 3 years.

\begin{figure}[t]
\centering
\includegraphics[width=0.9\textwidth]{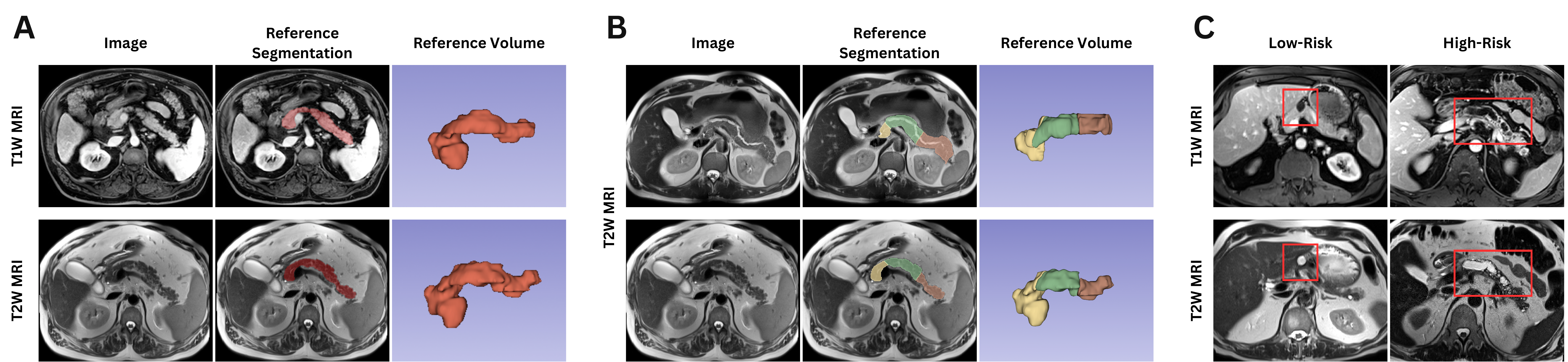}
\caption{Visualization of representative cases and segmentations. \textbf{(a)} Whole organ pancreas segmentation on T1W and T2W MRI. \textbf{(b)} Subregional pancreas segmentation on T2W MRI (head/body/tail). \textbf{(c)} Cases corresponding to binary (low/high) labels on T1W and T2W MRI, boxes indicate cyst location.}
\label{fig:medviz}
\end{figure}

The pancreas was manually segmented by expert radiologists. For the sub-regional setting, the volume was partitioned into three anatomical regions: the head/body boundary was demarcated by the medial border of the superior mesenteric vein, and the body/tail boundary was defined by splitting the remaining parenchyma along the middle of the longest transverse axis. The resulting masks are partially visualized in Fig.~\ref{fig:medviz}.

\begin{table*}[t]
\centering
\caption{Cohort breakdown by center. Age as mean [min,max], sex F/M/NA, class low-risk/high-risk. Both tracks use paired T1W/T2W; sub-regional counts are per region. Centers: NU = Northwestern University; NYU = New York University; MCF = Mayo Clinic Florida; EMC = Erasmus Medical Center; IU = Istanbul University; MCA = Mayo Clinic Arizona; AHN = Allegheny Health Network. $^{\ast}$Age was not recorded at AHN, so its per-center mean is unavailable and the totals average the six centers with recorded ages.}
\label{tab:data-breakdown}
\footnotesize
\renewcommand{\arraystretch}{1.10}
\setlength{\tabcolsep}{3pt}
\begin{tabular}{@{}l c c c c|c c c c@{}}
\hline
 & \multicolumn{4}{c|}{Whole-organ ($n{=}629$)} & \multicolumn{4}{c@{}}{Sub-regional ($n{=}616$)} \\
\cline{2-9}
Loc. & n & Age & Sex & Class & n & Age & Sex & Class \\
\hline
NU  & 185 & 63.7 [18,89] & 97/88/0 & 169/16 & 183 & 63.7 [18,89] & 96/87/0 & 167/16 \\
NYU & 150 & 62.7 [23,83] & 86/61/3 & 127/23 & 146 & 62.8 [23,83] & 82/61/3 & 123/23 \\
MCF & 130 & 65.7 [26,90] & 71/59/0 & 67/63 & 127 & 65.7 [26,90] & 69/58/0 & 64/63 \\
EMC & 72 & 50.6 [18,78] & 33/37/2 & 57/15 & 71 & 50.3 [18,78] & 33/36/2 & 56/15 \\
IU  & 62 & 63.3 [23,87] & 36/26/0 & 49/13 & 60 & 63.6 [23,87] & 35/25/0 & 47/13 \\
MCA & 15 & 69.3 [50,91] & 8/7/0 & 6/9 & 14 & 70.7 [60,91] & 7/7/0 & 5/9 \\
AHN & 15 & $^{\ast}$n/a & 11/4/0 & 12/3 & 15 & $^{\ast}$n/a & 11/4/0 & 12/3 \\
\hline
Tot. & 629 & 62.5 [18,91] & 342/282/5 & 487/142 & 616 & 62.5 [18,91] & 333/278/5 & 474/142 \\
\hline
\end{tabular}
\end{table*}


\section{Methodology}\label{sec:method}

We frame multimodal IPMN risk stratification as stacked
generalization~\cite{wolpert1992stacked,ting1997stacked}: single-modality
classifiers produce class probabilities that a level-1 stacking classifier fuses
into a final risk estimate. Let $\mathcal{D}=\{(\bx_i,y_i)\}_{i=1}^{N}$, where
$\bx_i=(\bx_i^{(s)})_{s=1}^{S}$ denotes the multi-stream input for patient~$i$.
Each stream-specific input $\bx_i^{(s)}$ represents one combination of imaging
representation, MRI sequence, and region of interest: the representation is
either radiomics features or a CNN image encoder, the sequence is T1W or T2W,
and the region of interest is either the whole pancreas or one sub-region from
the head/body/tail decomposition. Radiomics streams contribute engineered
feature vectors, whereas deep-learning streams contribute the corresponding
image crops or volumes. We study a \emph{binary} label $y_i\in\{0,1\}$,
corresponding to low- versus high-risk disease, and an \emph{ordinal 3-class}
label $y_i\in\{0,1,2\}$ ordered
$\textsf{no}<\textsf{low}<\textsf{high}$. Our pipeline is depicted in Fig~\ref{fig:cupmi_pip}.
\subsection{Base classifiers}\label{sec:base}
\paragraph{Radiomics.}
For each radiomics stream we train a shallow class-balanced random forest base
classifier ($100$ trees, maximum depth $5$). To control the high
feature-to-sample ratio, inputs are pruned in-fold (SelectKBest, top-$150$ by
ANOVA $F$) and $z$-scored, with all statistics estimated on the training fold
only. Thus the radiomics stream contributes out-of-fold class probabilities,
not selected features or fitted preprocessing parameters, to the stacking layer.

\paragraph{Deep learning.}
In parallel, each imaging stream is encoded by a 2.5D ImageNet-pretrained
ResNet-18 that emits per-stream class probabilities. As an end-to-end volumetric
comparator we additionally train a 3D DenseNet-121 from scratch on the $96^3$ ROI.
In the full-fusion configuration there are $S{=}16$ streams: eight radiomics and
eight 2.5D deep-learning streams over
$\{$whole, head, body, tail$\}\times\{$T1W,T2W$\}$. All base classifiers enter the
meta-level on equal footing.

\subsection{Regularization via Gaussian sampling}\label{sec:gmm}
\paragraph{Meta-features.}
The output of each base classifier is a probability vector $\mathbf{p}^{(i)}_s\in\Delta^2$;
we stack the per-class log-probabilities across streams,
\begin{equation}\label{eq:meta3}
  \boldsymbol{\phi}(\mathbf{p}^{(i)})
  = \bigl[\,\log \mathbf{p}^{(i)}_1,\;\ldots,\;\log \mathbf{p}^{(i)}_S\,\bigr]\in\R^{3S},
\end{equation}
with $\log$ applied component-wise after clipping to $[\varepsilon,1]$
($\varepsilon{=}10^{-6}$). For $S{=}16$ this yields a $48$-dimensional meta-vector;
the binary analogue stacks the two class log-probabilities per stream. The Gaussian augmentation is then fit in this
unconstrained log-probability space.
We employ this stacked representation rather than hand-engineered
interaction features (e.g.\ entropy-based confidence or pairwise-disagreement
statistics); such features hard-code a fixed set of interactions but do
not scale to the 16-stream setting.

\paragraph{cUPMI.}
On the training fold we fit one Gaussian per class with a \emph{shared pooled
covariance},
\begin{equation}\label{eq:gmm}
  q_c(\boldsymbol{\phi}) = \N\!\bigl(\boldsymbol{\phi}\mid\bmu_c,\bSigma\bigr),
  \quad
  \bmu_c = \tfrac{1}{n_c}\!\!\sum_{y_i=c}\!\boldsymbol{\phi}^{(i)},
  \quad
  \bSigma = \operatorname{Cov}\!\bigl(\{\boldsymbol{\phi}^{(i)}\}\bigr)+\lambda I,
\end{equation}
with ridge $\lambda{=}10^{-4}$. We then draw $n_{\text{synth}}=\rho\,n_{\text{real}}$
synthetic meta-features balanced across classes and appended them to the training set of the level-1 stacking classifier. The synthesis ratio $\rho$ is chosen by inner cross-validation, thus the sampler is never fit on scoring data. We utilize a shared covariance
and single component to mitigate overfitting concerns. Previously tested configurations include per-class and BIC-selected
multi-component mixtures as well as full versus diagonal covariance. In the
high-dimensional stacked log-probability space the added flexibility was not
reliable: replacing the shared-covariance single Gaussian with a BIC-selected
mixture reduced the XGBoost radiomics-only gain from $\Delta$QWK $+0.022$
($100\%$ of seeds) to $+0.004$ ($40\%$). 

\subsection{Tree combiner and evaluation}\label{sec:tree-combiner}
As the clinical ideal is ordinal risk stratification, we report macro one-vs-rest AUC and
adopt the quadratic-weighted Cohen's~$\kappa$ (QWK)~\cite{cohen1968qwk} as our
primary metric (due to larger two-step penalty).
Cross-track fusion is evaluated in a fold-locked manner: radiomics learners are
refit within each outer training fold, while deep-learning out-of-fold logits
are precomputed once under a single shared $5$-fold split. The two are
concatenated into Eq.~\ref{eq:meta3} only within that fixed split, so as to avoid scoring against a re-shuffled fold. The synthesis ratio $\rho$ for a given level-1 stacker is selected via inner
cross-validation within the training fold (i.e., never on the reporting fold). Furthermore we report
repeated stratified $5$-fold cross-validation (on five or more seeds) as mean$\pm$SD with patient-level paired-bootstrap
$95\%$ CIs.

Our binary configurations are calibrated with Platt scaling fit on out-of-fold
training logits~\cite{platt1999probabilistic}, and neural-network outputs with
temperature scaling fit on a held-out calibration split~\cite{guo2017calibration}. Calibration is
used only to assess probability reliability (Brier score, ECE, and reliability
curves), not to select the reported discriminative model. Full multiclass
calibration remains planned, though notably does not
underpin the QWK or macro-AUC results. At the level-1 stacking layer, we evaluate three combiners: (i) class-balanced RF ($200$ trees and maximum depth $4$), (ii)
an XGBoost classifier ($200$ shallow trees, maximum depth $3$, learning rate
$0.05$, subsampling and column subsampling both $0.8$), (iii) and an $L_2$-regularized
logistic regression stack trained as a baseline.
\section{Results}\label{sec:results}

We evaluate binary IPMN risk by AUC (which is calibration invariant) and 3-class ordinal risk
by QWK (with macro one-vs-rest AUC), reported as QWK~[macro-AUC]. Unless otherwise noted,
fold-locked 3-class runs use a fixed outer-fold partition ($n{=}616$; class balance
$154/320/142$) and $5$ seeds.

No single architecture dominated on either target (Table~\ref{tab:arch_compare}). The end-to-end 3D DenseNet-121 trails both the lighter 2.5D ResNet-18
and the radiomics stack (binary AUC $0.774$ vs.\ $0.784$ vs.\ $0.813$; QWK
$0.452$ vs.\ $0.497$ vs.\ $0.514$). The best binary AUC ($0.839$) and ordinal
QWK ($0.595$) correspond to the fused deep learning and radiomics paradigm, framing our contribution as one of fusion effectiveness and
combiner regularization.

\begin{table}[t]
\centering
\caption{Architecture and fusion comparison on the Cyst-X cohort. AUC is binary; ordinal performance is reported as QWK with macro-AUC in brackets. The bold row is the flagship fold-locked fused configuration.}
\label{tab:arch_compare}
\footnotesize
\setlength{\tabcolsep}{4pt}
\begin{tabular*}{\textwidth}{@{\extracolsep{\fill}}p{0.46\textwidth}ccc@{}}
\toprule
\textbf{Method} & \textbf{Init.} & \textbf{AUC} & \textbf{QWK [mAUC]} \\
\midrule
Radiomics, whole-organ $+$ $L_2$-LR stack
& -- & 0.813 & 0.514 [0.772] \\
2.5D ResNet-18~\cite{he2016resnet,deng2009imagenet}, T2W
& ImageNet & 0.784 & 0.497 [0.741] \\
3D DenseNet-121~\cite{huang2017densenet,yao2025radiomics_dl_ipmn}, T1W$+$T2W
& Scratch & 0.774 & 0.452 [0.714] \\
Fused DL$+$radiomics $L_2$-LR stack
& Mixed & 0.817 & 0.511 [0.758] \\
\textbf{Fused DL$+$radiomics RF stack}
& Mixed & \textbf{0.839} & \textbf{0.595 [0.800]} \\
\bottomrule
\end{tabular*}
\end{table}
cUPMI's effect appears proportional to combiner capacity. On the 2-stream whole-organ binary task
it increases AUC by $+0.015$ for the random forest and $+0.024$ for XGBoost (positive in
all seeds) but leaves the already-regularized $L_2$-logistic stack effectively
unchanged ($+0.0005$). Despite the linear stack's dominance in the 2-stream baseline, it fails to exploit cross-stream interaction with larger stream quantities. Under higher-modality settings, the
random forest overtakes LR even on binary AUC ($0.839$ vs.\ $0.817$).

\begin{figure}[t]
\centering
\includegraphics[width=\textwidth]{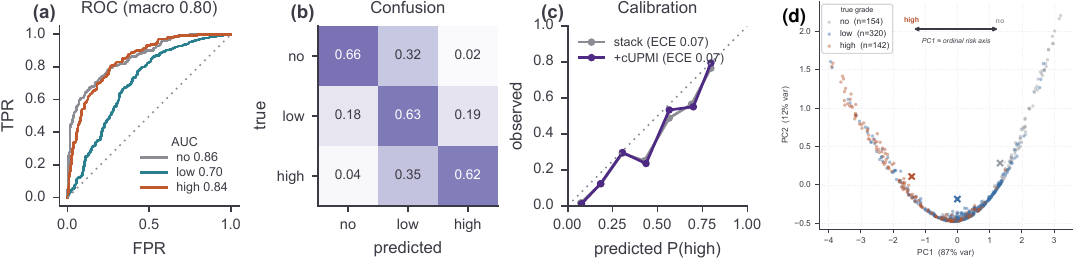}
\caption{Evaluation of the strongest full-fusion model (RF over fused DL+radiomics; 3-class, $n{=}616$).
\textbf{(a)} One-vs-rest ROC with per-class and macro AUC. \textbf{(b)} Row-normalized confusion: errors
concentrate on the \emph{adjacent} grade, with the costly two-step no$\leftrightarrow$high confusions near
zero ($0.02$/$0.04$). \textbf{(c)} Reliability of the predicted high-risk probability.
\textbf{(d)} PCA of the model's 3-class log-probability outputs (centroids $\times$): the ordinal axis is
recovered as PC1 ($87\%$ variance), with adjacent grades overlapping and the extremes ($\textsf{no}$,
$\textsf{high}$) separating---mirroring the confusion structure in (b).}
\label{fig:eval}
\end{figure}

When considering ordinal classification settings, cUPMI's benefit appears specific to high-capacity combiners
(Table~\ref{tab:ordinal_centerpiece}). On radiomics-only S8, cUPMI helps XGBoost
by $+0.022$ QWK in all seeds, whereas RF ($+0.004$) and LR ($+0.007$) show only
minimal gains. This pattern changes in the DL-Radiomics fusion setting: cUPMI fails to
improve RF stack performance ($\Delta$QWK $-0.003$) and slightly hurts LR ($-0.007$). XGBoost retains a negligible positive gain ($+0.005$). We believe the larger effect in the
fused setting derives from cross-track fusion itself, which lifts the tree
combiners from radiomics-only performance (RF $0.562{\to}0.595$, XGBoost
$0.521{\to}0.562$ QWK) while leaving LR flat. The strongest configuration is
therefore the fold-locked RF stack over fused DL$+$radiomics without cUPMI
(QWK $0.595$, macro-AUC $0.800$; $95\%$ CI $[0.54,0.64]$ from a $10{,}000$-resample
patient bootstrap). Errors appear concentrated on the adjacent grade, with near-zero
two-step confusions (Fig.~\ref{fig:eval}b).

\begin{table}[t]
\centering
\caption{Ordinal 3-class risk (fold-locked, $n{=}616$). Stack columns report QWK~[macro-AUC]; $\Delta$ columns report cUPMI-minus-stack QWK with the fraction of seeds improved. Best stack and clearest cUPMI gain are bold.}
\label{tab:ordinal_centerpiece}
\footnotesize
\setlength{\tabcolsep}{4pt}
\begin{tabular*}{\textwidth}{@{\extracolsep{\fill}}lcccc@{}}
\toprule
\textbf{Combiner} & \textbf{S8 stack} & \textbf{S8 $\Delta$} & \textbf{S8+DL stack} & \textbf{S8+DL $\Delta$} \\
\midrule
LR  & 0.514~[0.772] & $+0.007$ ($60\%$) & 0.511~[0.758] & $-0.007$ \\
RF  & 0.562~[0.782] & $+0.004$ ($40\%$) & \textbf{0.595~[0.800]} & $-0.003$ \\
XGB & 0.521~[0.768] & $\mathbf{+0.022}$ ($\mathbf{100\%}$) & 0.562~[0.787] & $+0.005$ ($80\%$) \\
\bottomrule
\end{tabular*}
\end{table}

\section{Discussion}\label{sec:discussion}
Level-1 combiners are a standard decision-level fusion tool in medical
imaging~\cite{huang2020fusion,stahlschmidt2022multimodal}. Because they are
trained on a small meta-feature set, higher-capacity tree techniques (e.g.,
RF~\cite{breiman2001rf} and XGBoost~\cite{chen2016xgboost}) are prone to
overfitting. cUPMI's effect is therefore a property of the combiner: $L_2$
stacks don't benefit while tree combiners do ($+0.015$ RF, $+0.024$ XGBoost
binary AUC, all seeds). By enriching the stacking classifier's input, cUPMI is
distinct from instance-space resampling such as SMOTE~\cite{chawla2002smote} and
from mixup~\cite{zhang2018mixup}. Its clearest gain (XGBoost, radiomics-only) is
modest in isolation, but the individually weak deep and sub-regional streams
(macro-AUC $0.66$--$0.74$), fused under a high-capacity combiner, give the
strongest configuration---exceeding every single architecture including the 3D
DenseNet-121, so anatomical decomposition complements rather than replaces
whole-organ context. The per-combiner $\Delta$QWK remains small relative to
patient-sampling variability ($95\%$ CIs include zero), and the minority-class
$N$ ($142$--$155$) nears the regime where the sampler becomes unreliable;
external validation and full $3$-class calibration are in progress.

\section{Conclusion}\label{sec:conclusion}

We present cUPMI, a class-conditional Gaussian regularizer for higher-capacity,
tree-based ensemble combiners. cUPMI provides its clearest benefit on ordinal
stacking tasks, particularly XGBoost in the 8-stream radiomics setting. We
additionally find that fusing whole-organ and sub-regional radiomics with
deep-learning streams (under fold-locked random forest) yields the strongest
overall model (QWK $0.595$, $95\%$ CI $[0.54,0.64]$; AUC $0.839$), surpassing
radiomics, 2.5D ResNet, and 3D DenseNet-121 on their own. External validation
and multiclass calibration remain as future work. Together, our results
suggest that careful regularization and information fusion can outperform
heavier architectures on small ordinal cohorts.

\bibliographystyle{plain}
\bibliography{ref}

\end{document}